# Improving Cardiovascular Disease Prediction Through Comparative Analysis of Machine Learning Models: A Case Study on Myocardial Infarction

Jonayet Miah
Department of Computer Science
University Of South Dakota
South Dakota, USA
Jonayet.miah@coyotes.usd.edu

Duc M Ca
Department of Economics
University of Tennessee
Knoxville TN, USA
ducminhcao1989@gmail.com

Md Abu Sayed
Professional Security Studies
New Jersey City University, Jersey City
New Jersey, USA
msayed@njcu.edu

Ehsanur Rashid Lipu
Department of Computer Science
Gannon University
New York, USA
Lipu001@gannon.edu

Fuad Mahmud
Department of Information Assurance and Cybersecurity, Gannon University, Pennsylvania, USA
mahmud002@gannon.edu

S M Yasir Arafat
Department of Mechanical Engineering
Lamar University
Texas, USA
sarafat@lamar.edu

*Abstract:* Cardiovascular disease remains a leading cause of mortality in the contemporary world. Its association with smoking, elevated blood pressure, and cholesterol levels underscores the significance of these risk factors. This study addresses the challenge of predicting myocardial illness, a formidable task in medical research. Accurate predictions are pivotal for refining healthcare strategies. This investigation conducts a comparative analysis of six distinct machine learning models: Logistic Regression, Support Vector Machine, Decision Tree, Bagging, XGBoost, and LightGBM. The attained outcomes exhibit promise, with accuracy rates as follows: Logistic Regression (81.00%), Support Vector Machine (75.01%), XGBoost (92.72%), LightGBM (90.60%), Decision Tree (82.30%), and Bagging (83.01%). Notably, XGBoost emerges as the top-performing model. These findings underscore its potential to enhance predictive precision for coronary infarction. As the prevalence of cardiovascular risk factors persists, incorporating advanced machine learning techniques holds the potential to refine proactive medical interventions.

## I. Introduction

The heart stands as the body's most essential organ, responsible for supplying blood to all organs and bones. If its function falters, crucial organs like the brain cease operation, leading to rapid fatality. Given its potential lethality, cardiovascular disease has garnered considerable attention in medical research. Diagnosing and treating cardiovascular issues pose intricate challenges, warranting automated predictions to enhance subsequent medical interventions. Cardiovascular disease prognosis generally hinges on the likelihood of its development, influenced by factors such as smoking, high cholesterol, family history, obesity, hypertension, and a sedentary lifestyle. Detecting these factors relies on assessing the patient's symptoms. Lifestyle changes, workplace stress, and poor dietary habits have led to a surge in heart-related disorders. Myocardial infarctions, commonly known as heart attacks, are globally prevalent. Diverse cardiac conditions affect a substantial population, with fatalities due to myocardial infarction on the rise.

Addressing the imperative to enhance prediction accuracy for myocardial infarction, this study delves into the domain of artificial intelligence and machine learning, harnessing their potential to aid in the identification of cardiac illness [1,2,3]. In an effort to shed light on the factors underpinning the development of heart disease, the research strives to pinpoint the crucial determinants by utilizing a myocardial dataset and established machine learning methodologies. The central focus of this investigation revolves around demonstrating the viability of machine learning techniques in gauging an individual's susceptibility to heart disease. By scrutinizing a range of machine learning models, the study endeavors to uncover which specific factors play pivotal roles in causing myocardial disease. A comparative analysis of six diverse machine learning models - namely, LightGBM, XGBoost, Logistic Regression, Bagging, Support Vector Machine, and Decision Tree - forms the cornerstone of this research, ultimately aiming to identify the most accurate predictor.

By evaluating the accuracy of each of these models in predicting myocardial disease, this paper seeks to elucidate their respective strengths and limitations in a practical clinical context. The results from the study provide valuable insights into the performance of each model, with Logistic Regression, Support Vector Machine, XGBoost, LightGBM, Decision Tree, and Bagging achieving accuracies of 81.00%, 75.01%,

92.72%, 90.60%, 82.30%, and 83.01% respectively. Notably, the analysis points toward LightGBM emerging as the most promising contender among the evaluated models.

## II. Literature Review

Yazdani et al. [1] In the pursuit of heart disease prediction, the researcher employed the weighted association rule mining algorithm, incorporating the significance scores of pertinent attributes. To enhance the credibility of the approach, validation was sought from cardiologists. Their expertise was instrumental in evaluating the importance scores of key attributes and the diagnostic rules employed. Leveraging the UCI open dataset, a well-established resource in cardiovascular research, a remarkably robust confidence score of 98% was achieved in the accurate prediction of heart disease.

Hassan et al. [2] In this article, the author makes predictions about the existence of cardiac problems by utilizing data sourced from the UCI compilation. For the purpose of forecasting heart disease, they assessed machine learning (ML) methodologies. In a departure from previous research trends that did not leverage the UCI dataset concerning cardiovascular conditions, this study introduced Gradient Boosted Tree (GBT) and Multilayer Perceptron (MLP) as predictive models for early detection of heart disease. The outcomes showcased the noteworthy performance of these two machine learning classifiers – GBT and MLP – both achieving a high accuracy rate of 95% in forecasting the onset of coronary heart disease. Notably, among the classifiers assessed, Random Forest (RF) emerged as the most accurate, boasting a classification accuracy of 96.28%. This success was complemented by a specificity of 0.9628 and a sensitivity of 0.9537.

Khan et al. [3], A comprehensive comparison of six distinct machine learning models was executed to predict myocardial disease, resulting in outcomes that were considered satisfactory. The array of machine learning models investigated encompassed LightGBM, XGBoost, Logistic Regression, Bagging, Support Vector Machine, and Decision Tree. These models achieved individual accuracy rates of 79.06%, 72.90%, 83.85%, 84.60%, 72.80%, and 82.01%, respectively. The study revealed that the LightGBM model exhibited superior performance compared to the others. Consequently, based on this observation, it can be inferred that among these six models, LightGBM demonstrated the most exceptional performance. Our findings imparted an optimistic outlook for the potential advancement of myocardial infarction treatment. However, before any commercial application, particularly within the healthcare sector, further investigation and study are necessary.

Kayyum et al. [4] This study involves the compilation of a dataset and the application of Machine Learning Algorithms for data classification. A total of 345 instances, each characterized by 26 attributes, were gathered. These instances were derived from patients afflicted by myocardial infarction, along with accompanying symptoms. The class attribute encompasses three distinct categories: distinctive, non-distinctive, and both. The dataset was subjected to training using the K-Fold Cross Validation Technique, employing three specific Machine Learning algorithms: Bagging, Logistic Regression, and Random Forest. As a result, the research managed to demonstrate accuracy rates of 93.913%, 93.6323%, and 91.0145% for the machine learning algorithms, respectively.

Islam et al. [5] In this prediction task, the accuracy of heart disease holds a pivotal significance. Taking this concern into account, the researchers conducted an examination of a dataset related to myocardial conditions, aiming to prognosticate myocardial infarction through the application of well-known Machine Learning algorithms, namely K-Means and Hierarchical Clustering. This study encompasses the assembly of data and the categorization of data through the utilization of Machine Learning Algorithms. The investigators amassed a dataset consisting of 345 instances, each featuring 26 attributes, sourced from various hospitals within Bangladesh. This dataset was drawn from patients afflicted by myocardial.

## III. METHODOLOGY

### 1. Data Collection

All data were gathered manually from diverse clinics and hospitals situated in Dhaka, Bangladesh. During the data collection process, particular emphasis was placed on capturing the recent health conditions of patients afflicted by heart failure. The team compiled the data from multiple clinics and hospitals, and the list of these sources is detailed in Table I. The dataset encompasses a total of 600 instances, each characterized by 13 distinct attributes, as presented in Table II. Within this dataset, information pertaining to specific patients is incorporated, including a class attribute categorized as either "Distinctive" or "Non-Distinctive." Notably, the dataset consists of 12 distinctive attributes and one non-distinctive attribute. This compilation encapsulates treatment records associated with 600 patients affected by heart failure, with each patient profile encompassing 13 distinct clinical traits. These records were gathered over the duration of the patients' treatment journey. The authors visually depicted the correlation matrix in Figure 2Table 1: The medical institutions involved in data collection.

| 1. | Square Hospital |
| 2. | National Heart Foundation Hospital |
| 3. | Crescent Hospital |
| 4. | Appolo Hospital |
| 5. | Farida Clinic |

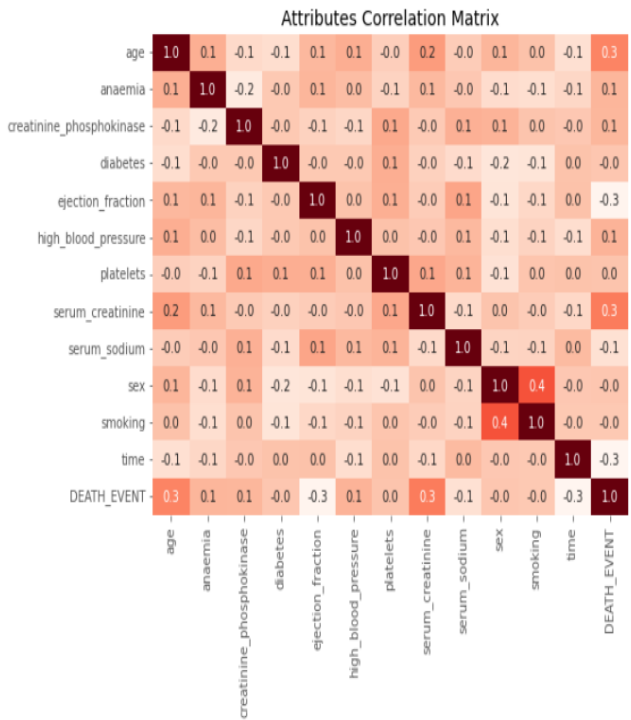

**Fig 2:** Correlation matrix between dataset attributes

## 2. Data Preprocessing & Filter

We utilized two unsupervised filters during the preprocessing phase using the renowned machine learning tool, Waikato Knowledge Analysis Environment (WEKA 3.8.3). To begin with, we removed the instances with absent items from the dataset, subsequently updating them. In this filtering procedure, we employed the mean, median, and mode to replace missing values in both qualitative and quantitative attributes. As the second step, we employed the Randomly Select filter, which effectively fills in the missing data points without causing substantial speed reductions. Additionally, we utilized the median function, which identifies the middle value within the dataset.

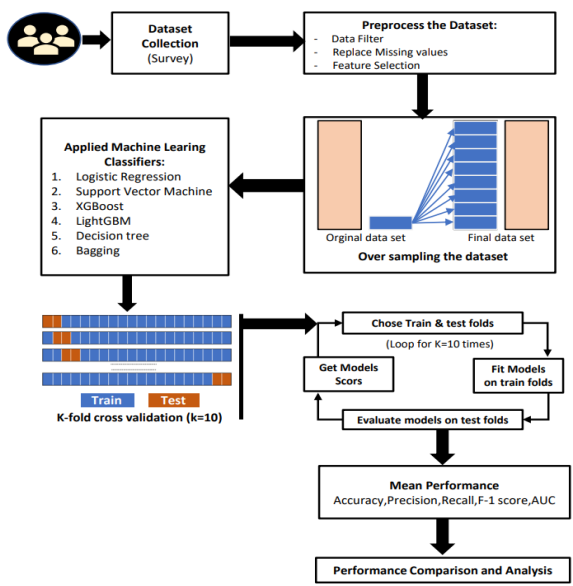

*Fig. 1: The overview of our study*

## 3. Feature Selection and Validation Technique

Rapid Miner: Rapid Miner finds utility in diverse domains, including academia, training, and research. In our application, this tool was utilized for various tasks, including data preprocessing, visualization of results, model validation, and process optimization. Acknowledged by Gartner as one of the most predictive analytical methods, Rapid Miner swiftly established itself as a leader in the advanced Magic Quadrant Systems Theory. The selection of an appropriate validation technique tailored to specific datasets is crucial. Opting for the most effective approach, the hold-out validation method involves allocating 80% of the dataset for training and 30% for testing. Implementing this method led to favorable outcomes. Metrics such as accuracy, sensitivity, specificity, and F1-Score were all computed using the implied confusion matrix. The presentation of a comprehensive assessment is facilitated through the use of bar graphs, effectively visualizing the performance indicators.

Table 2: Features List (Dataset attribute names)

| Attribute names | Attribute information |
|---|---|
| 1. Age: | The individual's age in years |
| 2. Sex | Man or woman (binary) |
| 3. Diabetes | Presence of diabetes in the patient (Boolean). |
| 4. Smoking Habit | Whether the individual was a smoker or not (Boolean). |
| 5. Anemia | Decrease in hemoglobin or red blood cells (Boolean) |
| 6. Blood Pressure | If the patient experiences elevated blood pressure (True/False). |
| 7. Creatinine phosphokinase (CPK) | The concentration of CPK enzyme in the blood (mcg/L) |
| 8. Time | Duration of observation (Days) |
| 9. Serum creatinine | Concentration of creatinine in the blood serum (measured in mg/dL) |
| 10. Serum sodium | The concentration of sodium in the bloodstream (measured in mEq/L) |
| 11. Ejection fraction | The proportion of blood expelled from the heart following each contraction expressed as a percentage. |
| 12. Platelets | The existence of blood platelets (platelet count per milliliter) |
| 13. Death event [Target] | At the time of the follow-up, if the patient had deceased (Boolean value). |

## Machine Learning Algorithms

After meticulous data preprocessing, rigorous training, and thoughtful categorization, a comprehensive array of machine-learning algorithms was employed in the analysis. These encompassed a diverse set of methodologies such as Logistic Regression, Support Vector Machine, XGBoost, LightGBM, Decision Tree, and Bagging. Following meticulous evaluation, the algorithm demonstrating the most remarkable performance was thoughtfully singled out. In a consolidated presentation, the outcomes of all these algorithms on the dataset are laid bare, offering a comprehensive view of their respective contributions, and enabling the identification of the highest-performing algorithm.

## Logistic Regression

In the pursuit of forecasting dependent categorical outcomes, a supervised learning approach known as logistic regression is harnessed. This methodology proves exceptionally valuable when dealing with vast datasets involving regression models. Through this algorithm, the likelihood of specific class probabilities is predicted based on pertinent dependent variables [14]. Mathematically encapsulated within the equation, $y = e^{(b0 + b1x)} / (1 + e^{(b0 + b1x)})$, 'x' represents the input value, 'y' signifies the anticipated outcome, 'b0' denotes the bias or intercept term, and 'b1' stands for the input coefficient.

The efficacy of logistic regression hinges on the Sigmoid function, adept at translating continuous outputs into probabilistic statements between 0 and 1. Elevating precision in this technique involves several pivotal steps: initial library importation, dataset visualization, handling null or missing values, data cleansing by removing extraneous elements, addressing outliers, defining independent and dependent variables, partitioning data into training and testing subsets, leveraging Ensemble and Boosting Algorithms, and engaging in Hyperparameter Tuning.

Furthermore, the study's primary focus involved the assessment of diverse predictive modeling classifications, which entailed the amalgamation of key metrics including accuracy, precision, recall, the F1-Score, and the area under the curve (AUC). The culmination of this comprehensive analysis is evident in Table III, which effectively summarizes the research findings. The comparative investigation extended across a range of machine learning classifiers, encompassing logistic regression, support vector machines, XGBoost, LightGBM, decision trees, and bagging. The evaluation process meticulously employed metrics such as accuracy, precision, recall, F1-Score, and AUC, with the outcomes of these evaluations thoughtfully compiled and visually presented in Figure 3.

## I. RESULT AND DISCUSSION

The investigation uncovered that upholding a standard platelet count enhances the likelihood of survival. Nonetheless, the connection between these factors exhibits minimal strength. Furthermore, sustaining a usual sodium level diminishes the fatality risk after heart failure. Conversely, elevated blood pressure amplifies the likelihood of demise following cardiac failure. The researchers observed that possessing a higher ejection fraction seems to decrease the peril of fatality after cardiac failure, although due to the limited sample size, it remains impractical to draw any inferences from extreme values. The association between the variables showcases limited strength, except for the link between gender and smoking.

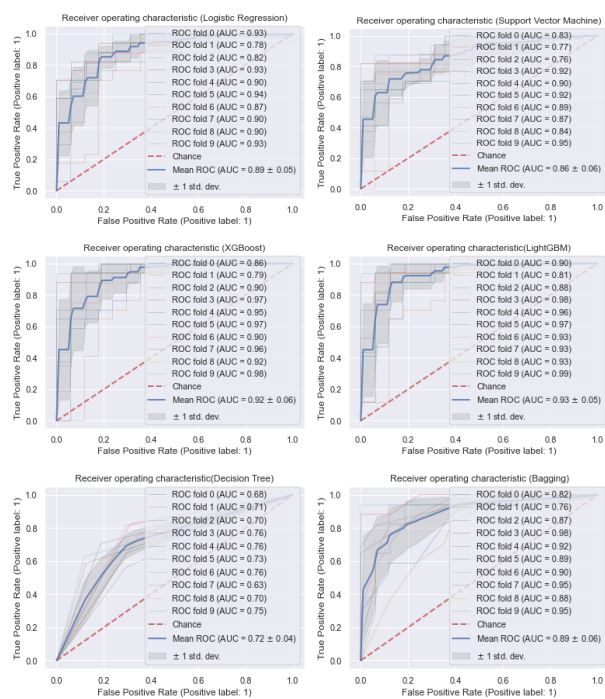

Fig 3: Area under curve output of a different kind of machine learning Algorithm

The interpretation of an Area under the Curve (AUC) value holds significance in gauging the performance of predictive models. An AUC of 0.5 signifies a scenario where no differential treatment is evident, indicating an inability to predict a patient's likelihood of cardiovascular disease based on the test outcome. Moving up the scale, an AUC between 0.7 and 0.8 indicates a commendable level of achievement, while a range of 0.8 to 0.9 reflects outstanding results. Remarkably, an AUC surpassing 0.9 reflects an exceptionally high level of performance from the system [13]. To visually capture the performance of specific machine learning models, a 10-fold cross-validation methodology is employed. This comprehensive visual representation encapsulates the comparative prowess of the various machine learning models, providing a clear perspective on their effectiveness, and aiding in informed decision-making. The estimation of a patient's vulnerability to cardiovascular disease is reliant on test outcomes. An Area under the Curve (AUC) ranging from 0.7 to 0.8 signifies a fitting level of achievement, while an AUC between 0.8 and 0.9 indicates excellent results. Moreover, an AUC exceeding 0.9 demonstrates exceptional system performance [13]. The evaluation of the designated machine learning models, facilitated by 10-fold cross-validation, is visually depicted in Figure 3. This graphical representation encompasses AUC plots along with average outcomes.

Table 3: Outcomes of different machine learning classifiers

| Machine Learning Model | Accuracy (%) | Precision (%) | Recall (%) | F1 Score (%) | AUC (%) |
|---|---|---|---|---|---|
| Logistic Regression | 86.0 | 83.00 | 78.6 | 78.90 | 88.0 |
| SVM | 75.0 | 72.04 | 81.1 | 75.80 | 87.0 |
| XGBoost | 92.7 | 87.00 | 81.0 | 83.90 | 91.0 |
| LightGBM | 90.6 | 86.30 | 81.9 | 84.00 | 93.9 |
| Decision Tree | 82.3 | 75.90 | 74.7 | 73.85 | 75.0 |
| Bagging | 83.0 | 86.96 | 74.5 | 74.00 | 86.0 |

Table 3 serves as a comprehensive repository of the outcomes derived from the application of six distinct machine learning classifiers, namely LightGBM, XGBoost, Logistic Regression, Bagging, Support Vector Machine, and Decision Tree. Notably, each classifier's prowess is evaluated through AUC scores, which are indicative of their performance in predicting the likelihood of the studied outcomes. Impressively, the classifiers attained AUC scores of 88.00%, 87.0%, 91.00%, 93.90%, 75.00%, and 86.00%, respectively. The results elucidate a discernible hierarchy in performance, where LightGBM and XGBoost emerge as the front-runners, demonstrating the highest AUC scores. This resounding success underscores their effectiveness in capturing and predicting the target variable. LightGBM garners attention not only for its remarkable AUC score but also for its commendable accuracy, reflecting its aptitude in making accurate predictions. In Chart 1, the comparative performance of the different machine learning models is vividly depicted, offering an intuitive visual aid for comprehending their relative strengths. This graphical representation enhances the clarity of the discussion and enables stakeholders to readily grasp the distinctions among the classifiers.

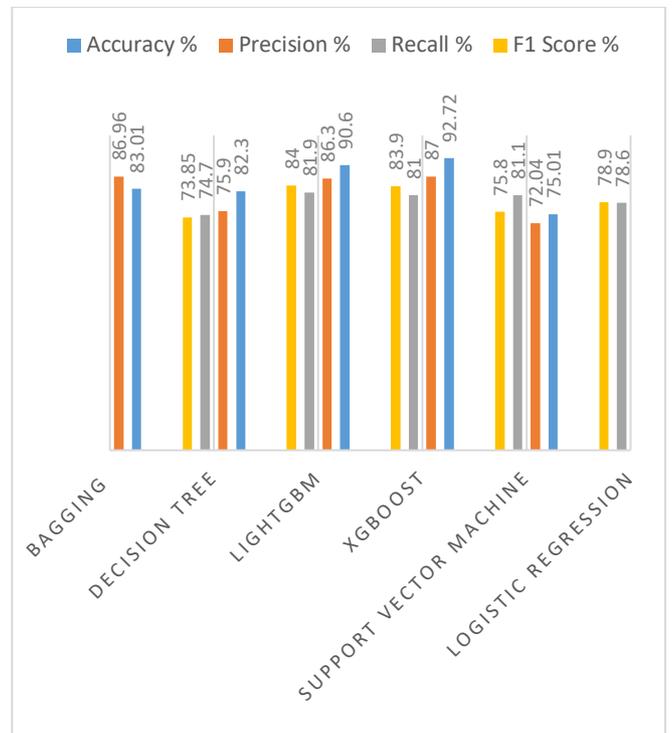

Chart 1. Analysis of the different machine learning models

V. CONCLUSION AND FUTURE WORK

With each passing day, the incidence of myocardial disease is on the rise. The positive aspect lies in the ability of machine learning to detect this ailment, offering a streamlined diagnostic approach. Our suggested model proves to be both effective and occasionally efficient in healthcare, enabling the early detection of this condition. The escalating incidence of myocardial disease has prompted a promising avenue for early detection through compact machine-learning models. Our proposed model stands as a potent tool in expediting diagnosis, ensuring timely intervention, and enhancing healthcare outcomes. Its efficacy extends to remote and underserved areas, where cost-effective diagnosis systems become invaluable. By leveraging machine learning's capacity to analyze extensive datasets of similar cases, our model offers reliable predictions. Amidst a plethora of myocardial disease prediction projects, our approach, notably XGBoost, achieved exceptional results, boasting 90.60% accuracy, 87.00% precision, 81.0% recall, 83.90% F1 score, and a remarkable 91.0% AUC. While promising, the widespread implementation of our model necessitates further research and development. Addressing challenges related to big data and exploring innovative solutions like blockchain can truly revolutionize cardiovascular health forecasting and pattern analysis.

In a landscape marked by an alarming rise in myocardial disease cases, the integration of machine learning emerges as a beacon of hope. Our proposed model demonstrates significant potential for early detection, offering a streamlined and effective diagnosis approach. These holds promise not only in established healthcare settings but also in underserved rural areas, ushering in a new era of accessibility. Through meticulous evaluation, our model, particularly exemplified by the impressive performance of XGBoost, underscores its accuracy and predictive power. The journey, however, is far from over. To fully harness the impact of our model, rigorous research, and innovation are imperative. The intersection of cutting-edge technology, informed research, and a commitment to healthcare advancement could propel cardiovascular health forecasting into uncharted territories. The path forward involves multifaceted exploration. Amidst the proliferation of machine learning models, our focus remains steadfast on refining and expanding our approach. We envision the utilization of distinct vocal feature-based datasets in conjunction with deep learning models, broadening the spectrum of myocardial disease detection. Tapping into advanced technologies, such as blockchain, could address data challenges and enhance the reliability of predictions. Moreover, the quest to predict cardiovascular disease recovery stands as an imminent challenge. Navigating this perilous territory demands the synthesis of predictive analytics, medical expertise, and technological innovation. As the healthcare paradigm transforms, our commitment to innovation and advancement remains unwavering, propelling us towards a future where proactive cardiovascular health management becomes a cornerstone of well-being.